# AR Visualization System for Ship Detection and Recognition Based on AI

Ziqi Ye

Fudan University
Shanghai, China
21210720276@m.fudan.edu.cn

Limin Huang

Chongqing University of Posts
and Telecommunications
Chongqing, China
2020210033@stu.cqupt.edu.cn

Yongji Wu

Chongqing University of Posts
and Telecommunications
Chongqing, China
2020210051@stu.cqupt.edu.cn

Min Hu*

Chongqing University of Posts
and Telecommunications
Chongqing, China
humin@cqupt.edu.cn

**ABSTRACT**

Augmented reality technology has been widely used in industrial design interaction, exhibition guide, information retrieval and other fields. The combination of artificial intelligence and augmented reality technology has also become a future development trend. This project is an AR visualization system for ship detection and recognition based on AI, which mainly includes three parts: artificial intelligence module, Unity development module and Hololens2-AR module. This project is based on R$^3$Det algorithm to complete the detection and recognition of ships in remote sensing images. The recognition rate of model detection trained on RTX 2080Ti can reach 96%. Then, the 3D model of the ship is obtained by ship categories and information and generated in the virtual scene. At the same time, voice module and UI interaction module are added. Finally, we completed the deployment of the project on Hololens2 through MRTK. The system realizes the fusion of computer vision and augmented reality technology, which maps the results of object detection to the AR field, and makes a brave step toward the future technological trend and intelligent application.

## 1 INTRODUCTION

The task of computer vision in artificial intelligence is to enable computers to understand the content of images in the same way as human eyes, and object detection has always been a hot topic in computer vision. Object detection technology is to determine the object to be detected in the image by algorithm, and at the same time mark the position of the object in the image and return the classification result. In recent years, the use of convolutional neural network knowledge has become popular in solving tasks such as object detection, which is superior to traditional detection algorithms in recognition accuracy and robustness. However, the mainstream interactive devices related to object detection are still traditional keyboard, mouse and touch screen, which have great limitations in object detection and human-computer interaction.

Augmented reality is a technology that superimposes virtual objects onto real-world scenes. The virtual object and the real world can achieve seamless superposition through the real-time calculation of the computer, so as to achieve the purpose of the fusion of virtual and real, to bring users a strong real feeling. The development of augmented reality and mixed reality technology can be applied to education, industry, medical and other industries. At present, most fields are still in the stage of exploration and development.

*e-mail: humin@cqupt.edu.cn

Microsoft HoloLens2 is the second generation of mixed reality glasses released by Microsoft in 2019. Its processor uses an Intel 32-bit CPU and a custom high-performance mixed reality computing unit (HPU). Compared with other AR devices, HoloLens2 uses a new interactive mode and 3D registration algorithm. Without additional auxiliary positioning devices, it can calculate the spatial position relationship between virtual objects and real scenes, so as to integrate the physical world and the digital world, and realize the virtual-real superposition, human-computer interaction and other technologies. Digital content can be displayed in the form of holograms. At the same time, holograms can be interacted with through gaze, gesture and voice.

By integrating the technologies of augmented reality and object detection and recognition, it can provide a new way of human-computer interaction, which provides theoretical and technical support for the problems existing in learning, training, visual display and other industries. This makes it possible to design and develop a deliverable solution.

Therefore, the project wants to combine object detection and recognition in computer vision with augmented reality. Finally, ships in remote sensing images can be detected and recognized, and then the recognized ships can be visualized by augmented reality technology. In addition, users can experience a range of human-computer interaction features through Hololens2.

## 2 METHODS

In this section, we define our task, complete the detection and recognition of ships in the image through the improved R3Det algorithm, and then build the scene and complete the function through Unity. Finally, the full functionality is presented through Hololens2.

### 2.1 Ship object detection network

The existing target detection algorithms can be roughly divided into Two categories: (1) Two-stage target detection algorithms based on Region Proposal, such as R-CNN[1], Faster R-CNN [2] and Mask R-CNN [3]. (2) One-stage object detection algorithms based on Regression, such as YOLO [4], SSD [5] and RRD [6].

Different from natural scene images, remote sensing images are generally images taken from the top perspective. Therefore, the direction of the object in the image is arbitrary, and there may be a dense arrangement of objects in the image. The horizontal bounding box in common target detection methods can not cover the target in any direction well in remote sensing images. However, the rotated bounding box with angle parameter can accurately surround the object with oblique angle and avoid the problems of large number of bounding boxes intersection caused by densely arranged objects. Figure 1 shows the difference between the two approaches.

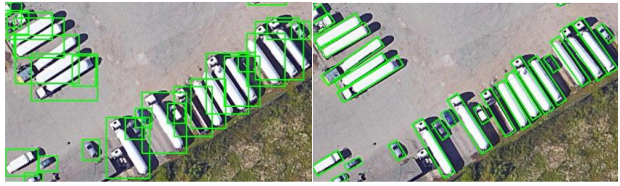

Figure 1: The difference between the horizontal bounding box object detection algorithm and the rotated box target detection algorithm

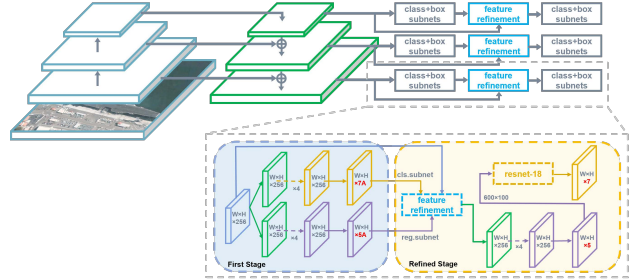

Figure 2: Improved R3Det network structure

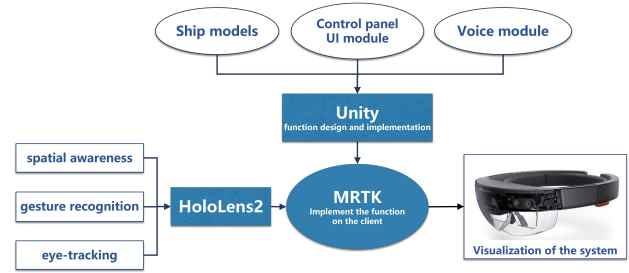

Figure 3: The technology roadmap of the system

According to the characteristics of remote sensing image objects, Yang et al. proposed a fast and accurate end-to-end rotating object detector named R$^3$Det [7]. R$^3$Det is a one-stage rotated object detector based on RetinaNet [8]. Rotated anchor boxes work better in scenarios where the targets are densely packed, while horizontal anchor boxes can achieve high recall with few numbers.

### 2.1.1 Improvement of Loss Function in R3Det Network

The existing Oriented Bounding Box (OBB) calculation methods basically introduce the offset angle parameter which is obtained by the distance loss on horizontal Bounding Box. However, it is insensitive to objects with large aspect ratios. This is because the distance loss reduces the Angle error of the directional bounding box and weakens the correlation with IoU.

PIoU Loss[9] can obtain more accurate results by using the pixel-level form. Therefore, this paper chooses PIoU Loss instead of Skew IoU Loss as the calculation method of IoU. The PIoU can be described as:

$$PIoU(b,b') = \frac{S_{b \cap b'}}{S_{b \cup b'}} \quad (1)$$

where the variable $b$ represents the predicted directional bounding box. The variable $b'$ represents the ground truth. For all positive samples $T$, the loss can be described as:

$$L_{PIoU} = \frac{-\sum_{(b,b') \in T} \ln PIoU(b,b')}{|T|} \quad (2)$$

According to equations (1) and (2), $PIoU$ is always greater than 0. There will never be a vanishing gradient problem. Therefore, it can be applied to directional bounding boxes and horizontal bounding boxes under normal conditions, and it can also be applied to bounding boxes without intersection.

### 2.1.2 Optimization of Classifier

The refining module of R$^3$Det only optimizes the bounding box regression part, so there is still much room for improvement in the object recognition of ships. Considering the complexity of the network and the feature that the residual network will not cause overfitting due to the high number of training iterations, ResNet [10] residual network will be used to replace the original category network of R$^3$Det in this paper. According to the analysis of ship data, the aspect ratio of most types of ships is close to 6:1.

Therefore, the object candidate box obtained by the R$^3$Det detection network is clipped, rotated and scaled to 3×600×100 image parameters, and then input into the ResNet-18 network that fits with it, finally completing the task of high-precision object positioning and prediction in this paper. The improved R$^3$Det network structure with PIoU and ResNet proposed in this paper is shown in Figure 2.

## 2.2 AR System Function Implementation in Unity

The technology roadmap of our system is shown in Figure 3. As you can see, Unity is a multi-platform integrated development tool. It can be distributed on Windows, Linux, Mac OS, iOS, Android, Web and so on. In addition, Unity can also be distributed to platforms like Hololens or Oculus using third-party toolkits, which can save developers a lot of time. Unity also features visual editing and dynamic previews, making it a great interactive experience for developers. Unity integrates the MonoDeveloper compilation platform and supports three scripting languages: JavaScript, Boo and C#. We use C# for programming in this project. Unity is a 3D game production engine in its infancy. Because of its powerful third-party resource package, it is widely used in augmented Reality, Virtual Reality, Mixed Reality and other fields. The system we designed is an augmented reality system developed by MRTK open source toolkit.

MRTK is an open source toolkit for mixed reality application development that can also be used in the augmented reality domain. With its cross-platform features, it can provide application development for Microsoft HoloLens, Windows Mixed Reality and other devices. MRTK for Microsoft HoloLens provides a modular build approach that helps reduce the size of projects. In addition, it can provide components for spatial interaction that can quickly migrate interspace object properties.

In this part, we completed the function design and implementation of AR system through Unity. First, we generated all the experimental images in 3D form in the scene. Secondly, the ship's position and category information are extracted according to the images output by the previous module. Thirdly, we search the model library for the corresponding model and generated a 3D model of the ship in Unity. In addition, we also added the voice module and the control panel UI module, which can make the system produce the speech introduction function and switch images, display data cards and other human-computer interaction functions.

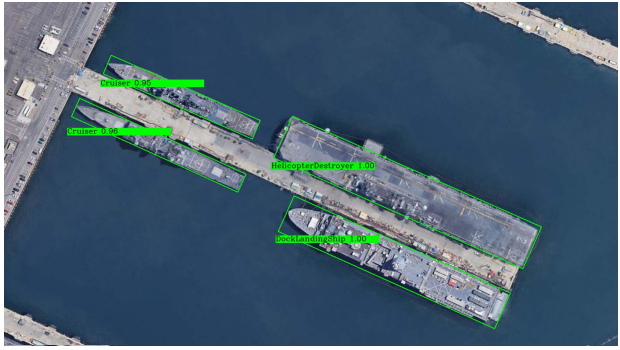

Figure 4: A example of our ship identification result

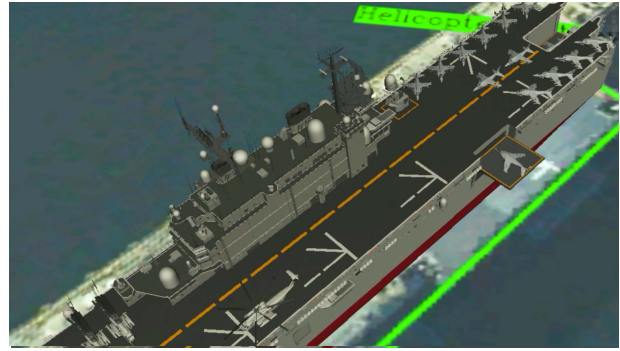

Figure 6: First perspective experience of our system

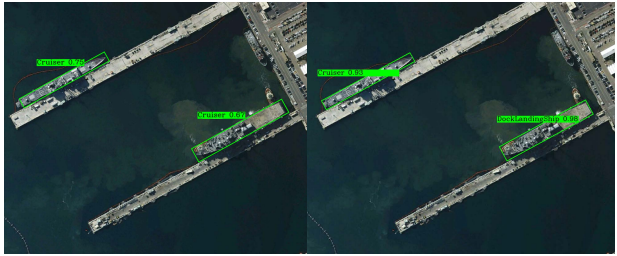

Figure 5: A identification results of R³Det algorithm and our algorithm

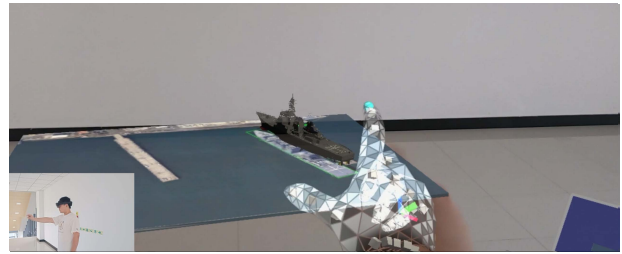

Figure 7: Practical experiments and third perspective

### 2.3 Client of the System: Hololens2

HoloLens2 is a mixed reality device that can accomplish mixed reality and augmented reality development tasks. HoloLens has the function of human understanding and environmental understanding. The human understanding part can realize hand tracking, eye tracking and speech recognition. This system mainly uses the functions of hand tracking and eye tracking. Hand tracking realizes direct interaction with the hologram by tracking the joint information of the user's hand, and eye tracking locates the position coordinates of the first visual Angle by tracking the user's eyeball. In the part of environment understanding, HoloLens2 provides six degrees of freedom tracking, which can be used for spatial location tracking and positioning to achieve position tracking worldwide. In addition, it can provide spatial mapping, which can determine the environmental grid in real time, and detect objects such as walls, floors, tables and chairs in the surrounding environment. The system is based on six degrees of freedom tracking technology to achieve user positioning.

In this part, We deployed the project on Hololens2 using MRTK. We deployed and applied the entire project on Hololens2 by utilizing spatial awareness, gesture recognition, and eye-tracking technologies. Users can experience the whole system by wearing the device.

## 3 RESULTS

In this section, we introduce the results from two sections: object detection and augmented reality.

### 3.1 Object Detection and Recognition of Ship

The experimental environment of this project is Linux system, CPU is AMD Ryzen 9 3900X, GPU is NVIDIA GeForce RTX 2080 Ti, and video memory is 11GB. CUDA 10.1 is used as the parallel computing architecture, Python 3.6.9 is used for programming, and PyTorch, NumPy, OpenCV and other Python packages are installed for development. PyTorch1.6.0 is an important deep learning framework of this project, which is used to build the neural network model.

The number of images in the dataset of this project is 2021, and the total number of samples is 4777. The target names for the seven categories of ships are: Aircraft Carriers, Helicopter Destroyers, cruisers, Dock Landing Ships, Destroyers, Frigates and Cargo Ships. In the stage of network training, the warm-up [11] strategy is adopted in the first five rounds. The weight attenuation factor is 0.0001, the momentum is 0.9, and the learning rate is 0.001. The ratio of training set to test set is 4:1, and we get a good result when epochs reach 100. In Table 1, we use mAP as evaluation index to show the detection and recognition effect of 7 kinds of ships. In Table 2, the ablation experiments are presented to demonstrate the effectiveness of our algorithm.

Table 1: The identification results of 7 ships

| AC | HD | Cr | DLS | Ds | Fr | Cs | **mAP** |
|---|---|---|---|---|---|---|---|
| 98.6% | 90.1% | 96.2% | 92.4% | 93.3% | 97.3% | 95.7% | **96.2%** |

Table 2: The results of ablation experiments

| Methods | mAP |
|---|---|
| R³Det | 78.6% |
| R³Det+ PIoU | 90.3% |
| R³Det + ResNet-18 | 85.7% |
| R³Det + ResNet-18 + PIoU (ours) | **96.2%** |

## 3.2 AR Visualization on Hololens2

This section introduces the AR visualization system of ship target detection results. Firstly, the user puts on the HoloLens2 device and opens the visual system project named "ship". Secondly, when the user's eye is fixed on the target ship, the system will voice the name of the ship. Thirdly, the user can point his finger at the object ship, and the system will trigger the ship's introduction card for display. In addition, users can switch more images to see the object detection results and AR visualization. Users can drag the switch function section to the right and click the corresponding image they want to view. The system will generate the identified image above the switch box and update the scene at the same time. The user's first perspective is shown in Figure 6, and the human-computer interaction action and the third perspective are shown in Figure 7.

## 4 CONCLUSION

The era of 5G has arrived, and the application of augmented reality technology in industrial design interaction, exhibition guide, information retrieval and other fields is gradually widespread. The combination of artificial intelligence and augmented reality technology has also become the future research and development trend. In this project, AR visualization and human-computer interaction of ships detected and recognized in remote sensing images are carried out by artificial intelligence algorithm. This is of great practical significance not only for the popularization of ocean ship knowledge but also for the deployment of maritime transportation.

As for the foreground, the functions and display effects of human-computer interaction can be improved in the future, and the model categories can be expanded at the same time. At present, this project has completed the identification, AR visualization and human-computer interaction functions of aircraft carriers, helicopter destroyers, cruisers and other ships. In the future, we can complete more types of ship identification and display, and even add more types of recognition display, such as aircraft, cars, iconic buildings, etc. In the future, this project can also be used as an independent terminal in the AR human-computer interaction experience area of the business district, or the exhibition area of the museum. This allows more people to experience and feel the wonderful visual effects of science and technology.

## REFERENCES

[1] Girshick R, Donahue J, Darrell T, et al. Region-Based Convolutional Networks for Accurate Object Detection and Segmentation[J]. IEEE Transactions on Pattern Analysis & Machine Intelligence, 2015, 38(1):142-158.

[2] Ren S, He K, Girshick R, et al. Faster R-CNN: Towards Real-Time Object Detection with Region Proposal Networks[J]. IEEE Transactions on Pattern Analysis & Machine Intelligence, 2017, 39(6):1137-1149.

[3] He K, Gkioxari G, Dollár P, et al. Mask R-CNN[C]// Proceedings of the IEEE International Conference on Computer Vision, 2017: 2961-2969.

[4] Redmon J, Divvala S, Girshick R, et al. You Only Look Once: Unified, real-time object detection[C]// Proceedings of the IEEE Conference on Computer Vision and Pattern Recognition, 2016: 779-788.

[5] Liu W, Anguelov D, Erhan D, et al. SSD: Single shot multibox detector[C]// European Conference on Computer Vision, 2016: 21-37.

[6] Liao M, Zhu Z, Shi B, et al. Rotation-Sensitive Regression for Oriented Scene Text Detection[C]// 2018 IEEE/CVF Conference on Computer Vision and Pattern Recognition. IEEE, 2018.

[7] Yang X, Liu Q, Yan J, et al. R3Det: Refined Single-Stage Detector with Feature Refinement for Rotating Object[J]. 2019.

[8] Lin T Y, Goyal P, Girshick R, et al. Focal Loss for Dense Object Detection[J]. IEEE Transactions on Pattern Analysis & Machine Intelligence, 2017, PP(99): 2999-3007.

[9] Chen Z, Chen K, Lin W, et al. PIoU Loss: Towards Accurate Oriented Object Detection in Complex Environments[C]// European Conference on Computer Vision (ECCV2020). 2020.

[10] He K, Zhang X, Ren S, et al. Deep residual learning for image recognition[C]// Proceedings of the IEEE Conference on Computer Vision and Pattern Recognition, 2016: 770-778.

[11] Gotmare A, Keskar N S, Xiong C, et al. A Closer Look at Deep Learning Heuristics: Learning rate restarts, Warmup and Distillation[J]. 2018.